\pdfoutput=1

\documentclass[11pt]{article}

\usepackage[preprint]{acl}

\usepackage{times}
\usepackage{latexsym}

\usepackage[T1]{fontenc}

\usepackage[utf8]{inputenc}

\usepackage{microtype}

\usepackage{inconsolata}

\usepackage{graphicx}

\usepackage{tabularx}
\usepackage{booktabs}
\usepackage{algorithm}
\usepackage{algpseudocode}
\usepackage{multirow}
\usepackage{rotating}
\usepackage{amsmath}
\usepackage{enumitem}

\usepackage{multirow}

\usepackage{varwidth}
\usepackage{calc} 
\algrenewcommand\algorithmicensure{%
  \makebox[\widthof{\textbf{Require:}}][l]{\textbf{Ensure:}}}

\algnewcommand\algorithmicforeach{\textbf{for each}}
\algdef{S}[FOR]{ForEach}[1]{\algorithmicforeach\ #1\ \algorithmicdo}

\usepackage{hyperref}

\title{Automating Easy Read Text Segmentation}

\author{Jesús Calleja$^{*1,2}$ \and Thierry Etchegoyhen$^{*1}$ \and David Ponce$^{1,2}$  \\
        $^1$ Fundación Vicomtech, Basque Research and Technology Alliance (BRTA) \\
        $^2$ University of the Basque Country UPV/EHU \\
        \texttt{\{jcalleja,tetchegoyhen,adponce\}@vicomtech.org} }

\begin{document}
\maketitle

\def\thefootnote{*}\footnotetext{Equal contribution.}\def\thefootnote{\arabic{footnote}}

\begin{abstract}
Easy Read text is one of the main forms of access to information for people with reading difficulties. One of the key characteristics of this type of text is the requirement to split sentences into smaller grammatical segments, to facilitate reading. Automated segmentation methods could foster the creation of Easy Read content, but their viability has yet to be addressed. In this work, we study novel methods for the task, leveraging masked and generative language models, along with constituent parsing. We conduct comprehensive automatic and human evaluations in three languages, analysing the strengths and weaknesses of the proposed alternatives, under scarce resource limitations. Our results highlight the viability of automated Easy Read text segmentation and remaining deficiencies compared to expert-driven human segmentation. 
\end{abstract}

\section{Introduction}

Being able to access Easy Read (ER) text, also known as Easy-to-read, is critical for large segments of the population, including people with cognitive disabilities or suffering from learning difficulties, among others.\footnote{The terms \textit{Easy-to-read}/\textit{Easy Read}, \textit{Easy Language} and \textit{Plain Language} are often used interchangeably, as they share a similar objective of facilitating the communication of information. There are however important differences between them, mainly that the target audience of ER content consists of people with reading difficulties, often with cognitive disabilities. ER text needs to be validated via focus groups, and the segmentation constraints are mainly meant to facilitate reading for this target group.} This type of text has two main characteristics. First, its content needs to consist of short sentences, using simple vocabulary and grammar as in text simplification, while also including an explanation of complex concepts in simpler terms as needed. Additionally, sentences need to be split into separate lines, following natural linguistic boundaries, to facilitate reading.\footnote{\url{https://www.inclusion-europe.eu/easy-to-read/}} 

Although several aspects of ER text automation have been studied \citep{readi-2020-tools,readi-2022-tools,readi-2024-tools}, to our knowledge automatic segmentation to meet ER readability requirements has not yet been explored in detail. The closest related domain where readability constraints require text segmentation along natural linguistic boundaries is subtitling.  Despite this similarity, subtitling features specific constraints, such as the number of characters per line or characters per second, which are tied to the audiovisual environment. For ER text, these constraints do not apply and there is a clear gap in our knowledge regarding optimal methods for text segmentation in this context.


In this work, we explore several methods for automated segmentation and evaluate them on expert-crafted ER content in three languages, namely Basque, English and Spanish. Considering the scarcity of ER resources, we focus our study on methods  able to generate segmentation hypotheses with either minimal or no training data. 
We design a scoring-based approach compatible with both pretrained constituency parsers and pretrained Masked Language Models (MLM). Additionally, we explore the use of pretrained generative Large Language Models (LLM), querying the models in zero-shot or few-shot fashion, as well as fine-tuning models on the limited available data.

We test the selected methods on datasets extracted from easy-to-read material collected from trustworthy sources, performing both automated and manual evaluations of the results. Our main results show that MLM-based scoring is a stable alternative across metrics, which may serve as a reasonable basis for ER text segmentation, although it still lags behind expert-crafted ER text.  

Our main contributions can be summarised as follows:

\begin{itemize}

    \item A first comprehensive assessment of automated segmentation for ER text generation.

    \item Novel approaches to ER text segmentation, including the use of generative LLMs under different modalities and a scoring method compatible with both constituency parsing and MLM modelling.

    \item Novel ER segmentation-centric datasets in Basque, English and Spanish, to support development in the field.\footnote{The datasets will be available under a CC-BY-NC-ND license at: \url{https://github.com/Vicomtech/EasyReadSeg}. As of this writing, note that efforts are still ongoing to secure sharing permissions for the content of all datasets.}

    \item A comprehensive evaluation in terms of automated metrics and human qualitative assessments over system segmentation hypotheses as well as human references.

\end{itemize}

\section{Related Work}
\label{sec:relw}

Aspects of Easy Read text generation are related to work on text simplification, as it involves standard lexical and grammatical simplification processes \cite{saggion2017automatic,alva-manchego-etal-2020-data,al2021automated}. Our work is more specifically related to Easy Read content generation constraints, including sentence segmentation, and automated text segmentation approaches. We describe each aspect in turn below.

\subsection{Easy Read}
\label{subsec:relw-er}

Easy Read is a text adaptation approach which aims at making information accessible to people with reading difficulties. A set of guidelines has been defined for ER document preparation across multiple languages,\footnote{See, e.g., \url{https://www.inclusion-europe.eu/easy-to-read-standards-guidelines/}}
covering text content, document layout, and validation of text comprehensibility by members of the target audience. The latter includes people with intellectual disabilities, prelingual hearing disabilities, aphasia, dyslexia or attention-deficit/hyperactivity disorder, among others. ER material can also be useful to anyone with reading difficulties, including people with limited knowledge of a language, for example. 

In terms of text, the guidelines include recommendations at different levels. As for lexical choice, simple words should be used, and concepts deemed too complex should be explained in simple terms. In terms of syntax, sentences should be simple and short, avoiding complex constructions. 
Finally, a sentence should fit into a single line wherever possible, but should otherwise be split into separate lines at natural linguistic boundaries, "where people would pause when reading out loud" as indicated in the Inclusion Europe English guidelines.\footnote{\textit{Op. cit.} footnote 4, pp. 17.} 



In terms of length, a standard recommendation is to use sentences that comprise between 5 and 15 words, to enhance readability. Although this requirement is not specified in all ER guidelines,\footnote{This constraint is explicit in the Spanish ER norm.} it is generally considered standard practice in the field, for languages like English or Spanish at least. 

These guidelines are meant to facilitate reading along several dimensions, with studies establishing empirical support for the use of easy-to-read texts, although some of the recommendations might deserve further investigation \cite{fajardo2014easy}. \citet{gonzalez2024empirical} point out the relative lack of studies to firmly establish empirical support for ER guidelines. 

\subsection{Text Segmentation}
\label{subsec:relw-seg}

To our knowledge, automated text segmentation has not been explored for ER text generation. As noted in the introduction, it is however an important aspect of subtitle generation, where subtitles should be segmented at naturally occurring linguistic boundaries to enhance readability, in addition to other subtitle-specific constraints. Subtitle segmentation has been shown to have an important impact on their readability \citep{doi:10.1080/15213269.2010.502873,doi:10.1080/0907676X.2012.722651}.

Several approaches have been explored for automated subtitle segmentation. Thus, \citet{10.1007/978-3-319-13623-3_24} trained Support Vector Machine and Linear Regression models over professionally-created subtitles to predict subtitle breaks. This approach was later improved with the use of Conditional Random Fields \citep{ALVAREZ201783}. Other approaches jointly learn the generation of subtitle breaks within machine translation models \cite{matusov-etal-2019-customizing,karakanta-etal-2020-42}, an integrated approach which cannot be extended to the direct segmentation of text. Alternatively, \citet{papi-etal-2022-dodging} proposed a multilingual segmenter which generates both text and breaks and may be trained on textual input only, or on joint text and audio data. 

Excepting the aforementioned integrated methods, most supervised approaches to subtitle segmentation could, in principle, translate to the ER use case, provided sufficient training data. Unfortunately, compared to subtitling, ER resources are scarce. To our knowledge, no ER datasets with segmented text are currently available, for any language. Furthermore, as shown in Section~\ref{sec:corpora}, the amount of ER data collectable from trustworthy sources is minimal to train machine learning models and, for most languages, simply non-existent. We will overcome these limitations by leveraging approaches to the task that rely on a limited number of samples, or unsupervised methods. Among the latter, we will notably adapt the  approach of \citet{ponce-etal-2023-unsupervised}, which performs subtitle segmentation by computing the MLM-based likelihood of punctuation marks as an approximation to natural linguistic boundaries.

\section{Methodology}
\label{sec:method}

The segmentation problem for ER text can be formulated as follows. Given a source sentence $x$, compute an ordered sequence of segments $[s_1, s_2, ..., s_n]$ such that the concatenation of all segments is identical to the source $x$, and all segments $s_i$ follow natural linguistic boundaries, ending at a position in the sentence where a reader would naturally pause. Although intuitively clear, defining what \textit{natural linguistic boundaries} or \textit{natural pauses} actually mean can be a challenge. To avoid the need for a more specific definition, we will consider expert-crafted text segmentation as our reference standard and perform a manual evaluation of these reference datasets in Section~\ref{sec:heval-res}. 

We describe in turn below the different approaches we selected for ER segmentation. Further details regarding the specific models for each language and scenario are presented in Section~\ref{subsec:models}.

\subsection{Scoring-based Segmentation}
\label{sec:scoring-seg}

We first devised a simple approach compatible with any model that can assign a score to segmentation candidates between words in a given sentence. The sentence is tokenised into words, using white-space tokenisation. We then extract segmentation candidates within a fixed window, with minimum and maximum number of words per segment. The candidates are then sorted according to their score, as assigned by a given scoring function. In cases where there is insufficient text for segmentation, the remaining text is selected as candidate. To avoid the pitfalls of selecting an early segmentation which may negatively affect the segmentation of the remainder of the input, we follow a beam-search approach and select the path with the best overall score, computed as the mean of all segments in the path. Appendix~\ref{sec:appendix-seg-algo} provides further details on the segmentation algorithm.

The score of a given segment can be computed with any chosen method. In the next sections, we describe our two main approaches to compute segmentation scores, respectively based on constituency parsing and masked language modelling.

\subsubsection{Constituency Scoring}
\label{sec:parse-seg}

Our first scoring method is based on constituency parsing, using a pretrained model. Under this approach, segmentation scores are computed between any two words as the distance between the leaves associated to these words in the constituency tree. The intuition in this case is that, as the distance between nodes increases, the likelier it is that a break will preserve constituency groups. Additionally, given that the parsing tokenisation pipeline could either introduce new tokens or modify existing ones, and considering that our segmentation candidates should strictly align with white-space boundaries within the original text, we keep track of valid segmentation candidates throughout the scoring process, discarding any candidates that do not meet this criterion.

We chose constituency parsing, instead of alternatives such as chunking, as it provides higher parsing granularity and thus a higher number of segmentation points. One drawback of this approach is the limited coverage across languages. 
Nonetheless, it is still worth including this type of approach, as grammatical constituency is in line with what could be typically understood as delimiting natural linguistic boundaries.

\subsubsection{MLM Scoring}
\label{sec:mlm-seg}

As our second scoring method, we adopted the unsupervised scoring approach to segmentation described in \citet{ponce-etal-2023-unsupervised}. We selected their best-performing variant, where segmentation candidates are assigned a score based on the MLM-predicted likelihood of a punctuation mark occurring in a mask inserted to the right of a given word. 

The main difference between our approach and theirs lies in the segmentation algorithm itself. Whereas they follow a greedy approach, selecting the best segmentation candidate within a segmentation window and proceeding recursively over the remainder of the text, we perform beam search to account for earlier segmentation choices that might not be optimal over the remaining text.

\subsection{Generative LLM Segmentation}
\label{sec:gen-lm}

As our third approach, we queried pre-trained generative LLMs to generate segmented text. We used three different variants of this approach, described below. One notable potential drawback of this approach is the lack of guarantee that the generated text will not be altered beyond the introduction of line breaks. We measure this specific aspect in our experiments.

\paragraph{Zero-shot.} We use simple prompts under a zero-shot approach with instruction-tuned LLMs and retrieve the model's answers as our segmentation hypotheses. Given well-established LLM prompt sensitivity \cite{lu-etal-2022-fantastically}, we experimented with the following variants, similar in the three languages:\footnote{See Appendix~\ref{sec:appendix-prompts} for details on the prompts in Basque and Spanish.}

\begin{itemize}
    \item[P1.] Split the provided sentence into separate segments, inserting cuts where people would pause when reading out loud.
    \item[P2.] Split the provided sentence into separate segments that follow natural grammatical constituent limits.
\end{itemize}

Prompt P1 formulates the problem as per the guidelines referenced in Section~\ref{subsec:relw-er}, whereas prompt P2 refers to natural grammatical boundaries, avoiding references to reading pauses, which might be less directly interpretable by the model.

Both prompts are complemented with the following text, to include additional constraints regarding segment length and preservation of the original content: "\textit{Each segment should have between 5 and 15 words. The content of the original sentence should be strictly maintained and no new information should be added}".

\paragraph{Few-shot} As a second variant, we performed few-shot prompting, providing the model with the same instructions as in our zero-shot approach and adding 5 examples of segmentation sampled from the development sets of the ER corpora, manually verified for correctness. As for zero-shot variants, one limitation of this approach is the lack of quality instruction-tuned models for most languages, to achieve proper segmentation. 

\paragraph{Fine-tuning} Finally, despite the relatively low amounts of training data, we fine-tuned base LLMs with LoRA \cite{hu2022lora}, as this approach may overcome some of the limitations of zero-shot and few-shot approaches, such as prompt instability, i.e., result variation depending on prompt formulation. Further details on the setup in this scenario are provided in Appendix~\ref{sec:appendix-prompts}.

\section{Experimental Setup}
\label{sec:setup}

In this section we describe the corpora we prepared for the task, the specific models selected for each approach and language, and our evaluation metrics.

\subsection{Corpora}
\label{sec:corpora}

Our corpora were collected from three separate websites which feature expert-crafted ER text that adhere to the guidelines in the field. For English, we used ER articles from Inclusion Europe.\footnote{\url{https://www.inclusion-europe.eu/category/etr/}} For Basque, we used articles from Irekia,\footnote{\url{https://www.irekia.euskadi.eus/lf/eu}} the transparency portal of the Basque Government, where a subset of the news are adapted for ER. Finally, for Spanish, we used both the Irekia website and articles from Plena Inclusión.\footnote{\url{https://www.plenainclusion.org/publicaciones/}}

We used in-house tools to scrape the data from the websites and perform boilerplate cleanup, notably removing enumerations presented via bullet points. We employed simple heuristics to identify segmentation, relying either on HTML elements whenever they permitted the identification of a line break, or sequences of uncased words in consecutive lines within text fields. All identified line breaks were then replaced with a segmentation marker and the text was split into separate sentences with scripts from the Moses toolkit \cite{koehn-etal-2007-moses}. Table~\ref{tab:corpora} describes the core corpora statistics, in terms of number of sentences or segments, average sentence or segment length, and number or percentage of line breaks.

\begin{table}[]
\centering
\begin{tabular}{llccc}
\toprule
                                   &         & EN & ES & EU \\ \midrule  
\multirow{3}{*}{\rotatebox{90}{All}} 
& \# sentences                                   & 4,441  & 14,353 & 10,956     \\ 
& \# breaks                          & 3,840  & 11,904  & 7,832     \\
& \% breaks                          & 86.47 & 82.94 &  71.49  \\ 
\midrule
\multirow{2}{*}{\parbox[c]{0pt}{\rotatebox{90}{\centering no seg}}} 
&  \# sentences        & 1,920   & 6,960  & 5,649    \\
& Avg. sent. len.   & 7.62  & 7.73  & 6.11      \\ \midrule

\multirow{4}{*}{\begin{sideways} w/ seg \end{sideways} } 
& \# sentences            & 2,521 & 7,393  & 5,307      \\
& Avg. \# breaks            & 1.52  & 1.61  & 1.48        \\
& Avg. sent. len.     & 14.98 & 19.19   & 14.62    \\
& Avg. seg. len.                         & 9.83  & 11.92    & 9.91    \\

\bottomrule
\end{tabular}
\caption{Corpora statistics.}
\label{tab:corpora}
\end{table}

In all three languages, the majority of the sentences are segmented, still with notable differences between the most segmented (English) and the least segmented (Basque). The exact reasons for these differences are difficult to determine, as they could be attributed to a difference in ER text adaptation style. One possible additional explanation is that in Basque, being an agglutinative language, there are simply fewer separate words in a sentence and therefore, fewer segmentation options. 

Sentences without segmentation are typically short across the board, which is expected since splitting those sentences would result in separate lines that are too short. Sentences with segmentation abide by the general guidelines overall, with less than 15 words per sentence and segments under 10 words, on average. Interestingly, although some guidelines recommend a maximum of two lines per sentence, the actual data indicate an average of over 1.5 breaks per sentence, i.e. a preponderance of sentences split into more than two lines. 


To train and evaluate the different segmentation approaches, we selected the subset of sentences with at least one segmentation break and partitioned the data as shown in Table~\ref{tab:partition}. Note that that we performed additional filtering after identifying remnant noise in the data, mainly improperly split sentences with text following a final punctuation. Our main criterion for the partition was to reserve as large a set as possible for testing, instead of opting for slightly more training data with less accurate testing.

\begin{table}[]
\centering
\begin{tabular}{lccc}
\toprule
& EN & ES & EU \\ \midrule
train & 358 & 3,974 & 2,249  \\
dev & 500 & 500 & 500  \\
test & 1,500 & 1,500 & 1,500  \\
\bottomrule
\end{tabular}
\caption{Data partition statistics (number of sentences)}
\label{tab:partition}
\end{table}

\subsection{Models}
\label{subsec:models}

\paragraph{Constituency Parsers.} We selected Stanza \citep{qi-etal-2020-stanza} as our main tool for the task, using default pipelines and models for English and Spanish. Since there is no default coverage for Basque in Stanza, we integrated  the Benepar EU model \citep{kitaev-klein-2018-constituency, kitaev-etal-2019-multilingual}, which is trained on a filtered and adapted version of the Basque Constituency Treebank \citep{seddah-etal-2013-overview}. We adapted the Stanza code to incorporate this model, using the Stanza preprocessing and tagging pipeline available for Basque.

\paragraph{Masked Language Models.} For English and Spanish, MLM scoring was computed with the multilingual BERT \cite{devlin-etal-2019-bert} model;\footnote{\url{https://huggingface.co/google-bert/bert-base-multilingual-uncased}} for Basque, we used the IXAmBERT model \cite{otegi2020conversational}. We used default parameters and the \textit{fill-mask} pipeline from the \textit{transformers} library.\footnote{\url{https://pypi.org/project/transformers/}}

\paragraph{Generative Language Models.} For our zero-shot and few-shot experiments, we selected two different models. For all three languages, we used the Llama2-7B instruction model \citep{touvron2023llama}.\footnote{\url{https://huggingface.co/meta-llama/Llama-2-7b-chat-hf}} To test a larger language model, we also used GPT-4 via the OpenAI API,\footnote{\url{https://pypi.org/project/openai/}} specifically model \textit{gpt-4-0125-preview}. Considering the lack of public information regarding their training, along with potential test data leakage issues \cite{balloccu-etal-2024-leak}, results with these models should be considered with all due caveats. For our fine-tuning experiments with LoRA, we used Llama2-7B in English and Spanish. For Basque, we selected the Latxa-7B model v1, a Llama2 model tuned on Basque corpora \cite{etxaniz-etal-2024-latxa}.\footnote{\url{https://huggingface.co/HiTZ/latxa-7b-v1}} 

\subsection{Evaluation Metrics}

Standard segmentation metrics such as F1 accuracy assume that the hypotheses and references are identical, except for segmentation markers. However, this constraint would not necessarily hold for generative LLMs, since they may alter the original text when generating segmentation hypotheses. To address a similar issue with models that may generate imperfect text in subtitling, along with segmentation hypotheses, \citet{karakanta-etal-2022-evaluating} proposed the Sigma metric, which computes the ratio of achieved BLEU \cite{papineni-etal-2002-bleu} over an approximated upper-bound BLEU score. We include results with this metric to compare all models, along with BLEU scores to measure the impact of imperfect text generation. All metrics results were computed with the EvalSubtitle tool.\footnote{\url{https://github.com/fyvo/EvalSubtitle}}

\subsection{Human Evaluation}

For our manual evaluations, we randomly sampled two subsets of 100 segmented sentences each: one sampled from the hypotheses of our overall best performing system, namely MLM; the other was sampled from the human segmentation references in the respective test sets for each language. Although human segmentation was produced by experts in the ER field, different segmentations are possible in general for a given sentence, and determining an optimal one can be subjective. Additionally, preliminary examination of the human references indicated cases of segmentation that did not adhere to the guidelines. We thus also aimed to measure the quality and consistency of the references in this evaluation. 

To evaluate system hypotheses, there were 3 evaluators per language; for human segmentation, there were 6, 4 and 3 evaluators for English, Spanish and Basque, respectively. All evaluators were unpaid volunteers, proficient in the corresponding language. The evaluators were not ER experts, but ER segmentation guidelines are relatively straightforward, and we estimated that the concepts involved, such as grammatical boundaries, could be assessed by untrained speakers of a given language.

For each sentence, the evaluators had to answer 5 questions regarding the quality of the segmentation:

\begin{enumerate}[noitemsep]
    \item Do sentence splits occur where people would naturally pause?	(disagree/mostly disagree/mostly agree/agree)	
    \item Do sentence splits occur along natural grammatical boundaries? (disagree/mostly disagree/mostly agree/agree)			
    \item Would you have split the sentence differently to facilitate reading?	(yes/no)
    \item Would you have used more splits to facilitate reading?	(yes/no)	
    \item Would you have used fewer splits to facilitate reading?	(yes/no)	
    
\end{enumerate}

More details on the evaluation protocol are provided in Appendix~\ref{sec:appendix-heval}.

\begin{table*}[ht]
\centering

\begin{tabular}{@{}lccccccccc@{}}
\toprule
                   & \multicolumn{3}{c}{EN}                                             & \multicolumn{3}{c}{ES}                                             & \multicolumn{3}{c}{EU}                                             \\ \toprule
                   & Sigma                & BLEU                 & F1                   & Sigma                & BLEU                 & F1                   & Sigma                & BLEU                 & F1                   \\ \midrule
MLM                & 82.88                & \textbf{100.00}      & \textbf{40.80}       & 85.57                & \textbf{100.00}      & \textbf{42.16}       & 83.57                & \textbf{100.00}      & \textbf{41.71}       \\
Stanza             & 80.35                & \textbf{100.00}      & 32.44             
& \textit{83.79}       & \textbf{100.00}      & 36.32                & \textit{71.27}       & \textbf{100.00}      & 27.19                \\ \midrule

Llama2-ZS-P1       & \textit{50.50}	& \textit{48.83}	& 11.44	& 71.54	& 24.55	& \textit{5.11}	& \textit{66.43}	& \textit{32.41}	& 1.89 \\
Llama2-ZS-P2       & 56.32	& 57.40	& 13.73	& \textit{63.33}	& \textit{21.61}	& 7.00	& 70.66	& 73.47	& \textit{1.26} \\

GPT4-ZS-P1         & \textbf{87.95}       & 77.08       & \textit{5.46}        & 86.88                & 93.27                & 7.20                 & 85.52                & 88.72                & 6.32                 \\
GPT4-ZS-P2         & 84.72                & 86.04                & 8.68                 & 87.23                & 95.62                & 8.68                 & \textbf{86.14}       & 81.97                & 7.53                 \\ \midrule
Llama2-FS-P1       & 80.05                & 93.82                & 7.35                 & 84.56                & 77.35       & 5.64        & 80.71                & 73.25       & 6.28                 \\
Llama2-FS-P2       & 79.85       & 94.11                & 7.11                 & 84.36                & 78.04                & 6.62                 & 80.40                & 80.97                & 4.89        \\
GPT4-FS-P1         & 85.66                & 98.28                & 5.53                 & 89.48                & 99.50                & 6.58                 & 84.34                & 99.35                & 18.04                \\
GPT4-FS-P2         & 85.84                & 97.99                & 6.37                 & \textbf{89.89}       & 99.51                & 6.45                 & 83.20                & 99.04                & 10.42                \\ \midrule
Llama2/Latxa-FT-P1 & 85.06                & 99.80                & 14.63                & 89.37                & 99.50                & 7.55                 & 86.02                & 99.78                & 12.63                \\
Llama2/Latxa-FT-P2 & 84.60                & 99.73                & 18.32                & 89.86                & 99.68                & 6.64                 & 86.10                & 99.58                & 15.17       \\

\bottomrule
\end{tabular}
\caption{Comparative results on the ER segmentation test sets. BLEU scores were computed over text without breaks to measure input text alteration. Fine-tuned models were all based on Llama2, the Latxa variant being used for Basque. The models are grouped according to their main characteristics: scoring-based (MLM and Stanza), zero-shot (ZS), few-shot (FS) and fine-tuned (FT). P1 and P2 indicate the two prompt variants described in Section~\ref{sec:gen-lm}. Best results per metric and language are shown in bold, worst results in italics.}
\label{table:auto-res}
\end{table*}

\section{Results}

\subsection{Automatic Evaluation}

The comparative results on the Sigma, F1 and BLEU (excluding breaks) metrics, for all languages and models, are shown in table~\ref{table:auto-res}.

Taking all metrics intro consideration, the best performing system overall was MLM-based segmentation, which achieved markedly better F1 scores in all languages, was competitive in Sigma, and achieved perfect BLEU scores, as expected for an approach that does not modify the input text. The Stanza variant was outperformed by MLM, with close Sigma results but significantly worse F1 scores, though still markedly better than all generative approaches. Stanza scores were lower in Basque overall, highlighting the dependency of this approach on the robustness of the available parsing tools for a given language. 

For all generative approaches, F1 scores were drastically lower than those achieved by the scoring-based methods, a consequence, in part at least, of their potential text-altering characteristics. As indicated by BLEU scores, input text modification could reach significant amounts, with scores around or below 80 for Llama2-FS in ES and EU. The most preserving approach in this respect were fine-tuning of Llama2/Latxa and GPT4-FS, in all languages, with scores close to or above 99, although even small degrees of input text alteration would still imply a revision and eventual post-editing of the output text, in actual practice.


In terms of Sigma, the highest scores were achieved by GPT4 variants and fine-tuned Llama2/Latxa models. Results with the latter indicate that even relatively small amounts of training data can result in Sigma scores comparable to those of models of a much larger size such as the GPT4 models (purportedly). 



The results were quite similar across language overall, with no notable deviation on metrics results per model variant. This was slightly unexpected, since all three languages are markedly different in terms of morphosyntax. It might be the case that even coarse-grained grammatical boundary identification is sufficient for the task, with most models capturing the core cases across languages.


Prompt variation did not have a significant impact in these experiments. This is also slightly surprising considering that one prompt refers to reading pauses and the other to grammatical boundaries. The common reference to splitting in the prompts might be the determining factor in comparable model behaviour, although a more  detailed analysis would be needed to clarify this matter. 

Finally, opting for zero-shot or few-shot did not markedly impact the GPT4 model on Sigma and F1, but the former led to the lowest results, by a large margin and across the board, for Llama2-7B. A zero-shot approach might be suboptimal for this type of smaller model on this task.


\subsection{Human Evaluation}
\label{sec:heval-res}

\begin{table*}[ht]

\begin{tabular}{@{}lccc@{}}
\toprule
                                 & \multicolumn{3}{c}{MLM Segmentation}                                             \\ \midrule
                                 & EN                        & ES                        & EU                      \\ \midrule  
Respects reading pauses?         & 59.33\% / 1.93 \small{[}0.59{]} & 59.67\% / 2.08 \small{[}0.42{]} & 57.33\% / 2.18 \small{[}0.24{]} \\
Respects grammatical boundaries? & 65.00\% / 2.14 \small{[}0.56{]}   & 68.00\% / 2.31 \small{[}0.50{]} & 63.00\% / 2.31 \small{[}0.28{]} \\
Alternate segmentation?          & 55.00\% \small{[}0.56{]}        & 54.00\% \small{[}0.51{]}        & 44.67\% \small{[}0.44{]}        \\
Use more breaks?                 & 20.33\% \small{[}0.53{]}        & 23.00\% \small{[}0.46{]}        & 17.00\% \small{[}0.29{]}        \\
Use fewer breaks?                & 3.33\% \small{[}0.69{]}        & 2.00\% \small{[0.32]}              & 7.67\% \small{[0.25]}              \\ \midrule
                                 & \multicolumn{3}{c}{Human Segmentation}                                              \\ \midrule    
                                 & EN                        & ES                        & EU                     \\ \midrule   
Respects reading pauses?         & 70.20\% / 2.36 \small{[}0.27{]} & 69.75\% / 2.49 \small{[}0.12{]} & 51.00\% / 2.10 \small{[}0.23{]} \\
Respects grammatical boundaries? & 80.20\% / 2.55 \small{[}0.33{]} & 78.50\% / 2.68 \small{[}0.06{]} & 72.00\% / 2.47 \small{[}0.20{]} \\
Alternate segmentation?          & 33.20\% \small{[}0.32{]}        & 34.25\% \small{[}0.29{]}        & 50.00\% \small{[}0.39{]}        \\
Use more breaks?                 & 9.40\% \small{[}0.21{]}        & 18.75\% \small{[}0.21{]}        & 24.00\% \small{[}0.13{]}        \\
Use fewer breaks?                & 4.40\% \small{[}0.26{]}        & 6.50\% \small{[}0.07{]}        & 6.00\% \small{[}0.12{]}  \\
\bottomrule
\end{tabular}

\caption{Human evaluation results in English, Spanish and Basque, over segmentation samples from the MLM system and human experts. Answers to the first two questions are the percentage of \textit{totally agree} answers and the average score on a [0-3] scale. For the other questions, we report the percentage of Yes. Krippendorf's alpha inter-rater agreement is indicated between brackets.}
\label{table:human-res}

\end{table*}

For the human evaluation, we selected the optimal segmentation system from the results of the previous section namely the scoring-based MLM variant. The results are shown in Table~\ref{table:human-res} and segmentation examples are provided in Appendix~\ref{sec:appendix-examples}.



As a first notable result, automated segmentation achieved lower scores than human segmentation overall for English and Spanish, outperformed by at least 10 percentage points in most cases. The system hypotheses were still valued positively overall, at 65\% on grammatical boundaries and around 45\% viewed as needing alternate segmentation. Note that even human reference segmentation was viewed as possibly requiring different segmentation in at least 33\% of the cases, which indicates the difficult and partly subjective nature of the task.

For Basque, surprisingly, the human reference breaks were viewed as less respectful of reading pauses, although still more grammatical overall. Alternate segmentation was also suggested in more cases for the human references. This highlights the difficulties of the task for human ER content creators as well, as  text segmentation can be perceived as suboptimal overall due to individual reading preferences. 




Both automatic and human segmentation received harsher scores in terms of reading pauses than grammatical boundaries. This could be due to the former being more subjective or dependent on individual readers than the latter. Inter-rater agreement was not markedly different though, with moderate agreement overall in both cases. This might thus simply indicate that the concept of natural reading pauses was more difficult to translate into proper segmentation.


Inter-rater agreement was higher on MLM segmentation overall, except for Basque where it was on a par or slightly higher than agreement on human references. This would tend to show that it was relatively easier to discriminate between correct and incorrect automated segmentation, with more salient errors when the automated system failed to generate a correct segmentation.


Finally, under-splitting was significantly more prevalent than over-splitting for both human and automated segmentation. This effect might be partially due to the evaluators not suffering from reading disabilities, thus differing from what focus ER groups would value as optimal to facilitate reading. We leave this type of analysis for future work.

\section{Error Analysis}
\label{sec:err_analysis}


\paragraph{Segmentation Window.} To assess the impact of modelling the guidelines recommendations as a fixed window, we performed a grid search by varying the min and max parameters of the scoring-based models. Overall, the selected 5-15 window (F1: 40.80) was optimal in English, closely followed by the 1-10 window at 40.14. For Spanish, 1-10 achieved 45.22 and 5-10 yielded 44.09, both outperforming our default (42.16). For Basque, 5-15 at 41.71 was outperformed by 5-10 at 43.89. The complete results can be found in Appendix~\ref{sec:appendix-grid}. 
        
Although improvements could have been achieved with different windows, these would have been somewhat marginal. Nonetheless, segmentation references outside our selected window would result in systematic errors for the scoring-based approach. There were 0.29\%, 0.19\% and 0.03\% cases of segments over 15 words, for EN, ES and EU, respectively, of minimal impact. There were however 26.34\%, 14.49\% and 24.86\% cases of segments under 5 words, for EN, ES and EU, respectively. These cases contributed significantly to the errors of the scoring-based variants.



Considering the relatively lax minimum number of words per segment in the reference sets, a possible solution would be to use a predefined segmentation window as a soft constraint rather than a hard limit on the segmentation candidates. 



\paragraph{Ungrammatical Segmentation.} This type of error is the most damaging for the task, as it directly impacts text readability. Although it would be difficult to uncover all failure patterns, we manually examined all cases in English where human evaluators assigned the worst score over both grammaticality and reading pause, in the human evaluation, which amounted to 36 examples.

Wrongly segmented complex named entities accounted for 25\% of these errors, with cases such as: \textit{Now, Thibeau works at Antwerp <seg> Management School as a researcher.} 

Another pattern which also accounted for 25\% of the cases were breaks after words that can act as a function word or as an independent word, e.g.: \textit{One person working in the institution told her that <seg> she could live on her own.}. 

One additional pattern emerged, though only in 7\% of the cases: split nominal coordination, influenced by the fact that a segment before a coordination can typically act in isolation, as in: \textit{Policies can be a set of rules <seg> or guidelines to follow in or to achieve a specific goal.} 

NER preprocessing, heuristics, or fine-tuning the pretrained model on segmentation data could be alternatives worth exploring for these types of errors, at the cost of a more complex pipeline.






\section{Conclusion}

We described a comprehensive evaluation of automated segmentation for Easy Read text generation, a critical component of accessible content for people with reading difficulties.

We explored a wide range of methods, including scoring-based segmentation with masked language models or constituency parsing, and generative language models via zero-shot or few-shot prompting, and fine-tuning. Although limited ER resources are available overall, we collected data that supported minimal training and relatively large test sets in Basque, English and Spanish.

In terms of metrics, the unsupervised MLM scoring approach proved optimal overall, although generative models could achieved higher Sigma scores. The main deficiency of the latter was the generation of input altering text and low F1 scores for all variants. Fine-tuned LLMs generated the least amount of imperfect text, among generative variants, and might be a worthy alternative for the task. 

Our human evaluations targeted both MLM-based and human segmentation, with the former lagging behind the latter in most cases, although the results were respectable overall across languages. We also identified segmentation error patterns from the MLM approach, which may provide research paths for future improvements.











\section{Limitations}

The evaluators that took part in our evaluation were not experts in the ER field, i.e. not trained to assess the readability of a given segmented sentence. Mitigating this limitation is the fact that the segmentation guidelines are relatively straightforward and the concepts involved, such as grammatical boundaries, can be assessed by untrained speakers of a given language. Nonetheless, ER text is specifically meant for people with reading difficulties, and an assessment of ER segmentation would gain from being performed by both experts in the field and focus groups. We leave this type of experimentation, which would involve more complex experimental protocols, for future studies. 

\section{Ethical Considerations}

Automated methods for text segmentation are meant to facilitate reading for people with reading difficulties, including people with cognitive disabilities. These methods are experimental, with no guarantee that the output they produce will be error free or meet Easy Read standards. All automatically generated output should undergo human revision and be corrected as needed.

\section*{Acknowledgments}

We wish to thank the anonymous ARR reviewers for their helpful comments, and the participants in our manual evaluations for their time and contributions. This work was partially supported by the Department of Economic Development and Competitiveness of the Basque Government (Spri Group) via funding for the IRAZ project (ZL-2024/00570).

\bibliography{anthology,custom}

\onecolumn
\newpage
\appendix
\section{Scoring Algorithm}
\label{sec:appendix-seg-algo}

The scoring algorithm described in Section~\ref{sec:appendix-seg-algo} is presented in more detail in Algorithm~\ref{algo:dynseg}. 

Each sentence is tokenised in to an array of words, via white-space tokenisation, and passed to a scoring function, either one of the methods described in Section~\ref{sec:scoring-seg} in our experiments. The scoring function performs its own pre- and post-processing of the array of words, and returns an array of \textit{scores} matching the number of words.

The $beam\_search$ function receives the initial lists of words and segmentation scores. Both lists are aligned to have their indexes match, by including a score after the final word. The function applies a standard beam search to keep the number of candidates to the maximum indicated as $beam\_width$, effectively reducing the search space to the top $k$ best scoring candidates. 

The $get\_segmentation\_candidates$ function, receives a list of words and its associated segmentation scores. The parameters $min\_words$ and $max\_words$ indicate predefined minimum and maximum number of words per segment, respectively. The function creates segmentation candidates within the window defined by these two parameters, where each candidate is assigned the score at its segmentation point, the words up to the segmentation point, and the remaining words and scores from this segmentation point on, if any. If the number of words to segment contains fewer than $min\_words$, the entire list of words is returned as a single segment and is assigned a penalty to decrease the likelihood of final segments with fewer than the required number of words.

\begin{algorithm}    
\begin{algorithmic}
\Function{$get\_segmentation\_candidates$}{$words, scores$}
\If {$length(words) < min\_words$}
\State $\textbf{return}$ $\{words: words, score: penalty, remaining\_words: \emptyset, remaining\_scores: \emptyset\}$
\EndIf

\ForEach{$index$ $\textbf{in}$ $range(length(words))$}
\If {$ min\_words > index < max\_words$}
\State $c \gets new\_candidate()$
\State $c.score = scores[index]$
\State $c.words = words[:index]$
\State $c.remaining\_words = words[index+1: ]$
\State $c.remaining\_scores = scores[index+1: ]$
\State $candidates.append(c)$

\EndIf
\EndFor 

\State $\textbf{return}$ $candidates$
\EndFunction
\end{algorithmic}

\begin{algorithmic}
\Function{$beam\_search$}{$words, scores, beam\_width$}

\State $candidates \gets $ $get\_segmentation\_candidates(words, scores)$
\While{$true$}
    \State $new\_candidates \gets $ [ ]
    \ForEach{$c$ $\textbf{in}$ $candidates$}
    \State $sc \gets get\_segmentation\_candidates(c.remaining\_words, c.remaining\_scores)$
    \State $new\_candidates.append(sc)$
    \EndFor
    \State $sort\_by\_score(new\_candidates)$
    \State $candidates \gets new\_candidates[:beam\_width]$
    \State $best \gets candidates[0]$
    \If{$best.remaining\_words = \emptyset$}
    \State $\textbf{return}$ $best$
    \EndIf
\EndWhile
\EndFunction
\end{algorithmic}
\caption{Scoring-based Segmentation}\label{alg:cap}
\label{algo:dynseg}
\end{algorithm}

\section{Training and Inference}
\label{sec:appendix-prompts}

Fine-tuning was performed with LoRA \cite{hu2022lora}, loading the base model in 8 bits, with the following parameters: r=8, alpha=16, dropout=0.0; targeted layers were Query and Value. 

Other training parameters were as follows: warmup\_steps=100; max\_steps=10000; optimizer='adamw\_torch'; batch\_size=4; gradient\_accumulation\_steps=17; learning\_rate=3e-4; max\_grad\_norm=0.3; weight\_decay=0.01. 

For generation, we used default parameters, namely: max\_new\_tokens=256, temperature=0.1, top\_p=0.9.

Each training sample was provided in the following format, where $\alpha$ is the full sentence and $\beta$ its corresponding segmented sentence:

\begin{itemize}

    \itemsep0em 

    \item[] INSTRUCTION: Split the provided sentence into separate segments, inserting cuts where people would pause when reading out loud. Each segment should have between 5 and 15 words. The content of the original sentence should be strictly maintained and no new information should be added.

    \item[] INPUT: $\alpha$ 

    \item[] RESPONSE: $\beta$    

\end{itemize}

The following text was prepended to each sample prompt, in the corresponding language: 

\begin{itemize}

    \item \textit{English}: Below is an instruction that describes a task, paired with an input that provides further context. Write a response that appropriately completes the request.

    \item \textit{Spanish}: A continuación se muestra una instrucción que describe una tarea, junto con una entrada que proporciona más contexto. Escriba una respuesta que complete adecuadamente la solicitud.

    \item \textit{Basque}: Jarraian, ataza bat deskribatzen duen jarraibide bat agertzen da, testuinguru gehiago ematen duen sarrera batekin batera. Idatzi eskaera behar bezala osatuko duen erantzun bat.
\end{itemize}

The following prompts were used for each language:

\begin{itemize}

    \item Prompt P1:

    \begin{itemize}
        \item \textit{English}: Split the provided sentence into separate segments, inserting cuts where people would pause when reading out loud. Each segment should have between 5 and 15 words. The content of the original sentence should be strictly maintained and no new information should be added.

        \item  \textit{Spanish}: Parte la frase proporcionada en diferentes segmentos, insertando cortes donde las personas se pararían al leerla. Cada segmento debería tener entre 5 y 15 palabras. El contenido de la frase original debería manterse estríctamente y no se debería añadir información nueva.

        \item \textit{Basque}: Zatitu emandako esaldia segmentu desberdinetan, jendea irakurtzean geldituko litzatekeen lekuetan moztuz. Segmentu bakoitzak 5 eta 15 arteko hitz kopurua izan beharko luke. Esaldiaren edukia zorrozki mantendu beharko litzateke eta ez litzateke informazio berririk gehitu beharko.
        
    \end{itemize}

    \item Prompt P2:  

    \begin{itemize}
        \item \textit{English}: Split the provided sentence into separate segments that follow natural grammatical boundaries. Each segment should have between 5 and 15 words. The content of the original sentence should be strictly maintained and no new information should be added.

        \item \textit{Spanish}: Parte la frase proporcionada en diferentes segmentos que respeten fronteras lingüísticas naturales. Cada segmento debería tener entre 5 y 15 palabras. El contenido de la frase original debería manterse estríctamente y no se debería añadir información nueva.

        \item \textit{Basque}: Emandako esaldia muga linguistiko naturalak errespetatzen dituen segmentuetan zatitu. Segmentu bakoitzak 5 eta 15 arteko hitz kopurua izan beharko luke. Esaldiaren edukia zorrozki mantendu beharko litzateke eta ez litzateke informazio berririk gehitu beharko.
    \end{itemize}

\end{itemize}

Basque and English models were trained on 1 L40 with 48GB of RAM; Spanish models on 1 L40S with 48GB of RAM. Checkpoints were saved every 30 steps. The best checkpoints were achieved at steps 180 and 210 for prompts P1 and P2, respectively, for Basque; 330 and 390 for Spanish; and step 90 for both prompts in English.

For few-shot and zero-shot, we used the default Llama2 template, using the system role for the instruction and the user role for the sentence to be segmented. For few-shot prompting, 5 examples of segmentation were shown after the instruction, using the user and assistant roles for the input and output examples.

\section{Grid Search}
\label{sec:appendix-grid}

\begin{figure}[htbp]
    \centering
\includegraphics[scale=0.43]{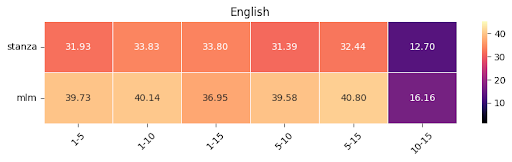} 
    \includegraphics[scale=0.43]{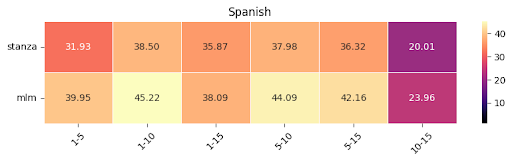}
    \includegraphics[scale=0.43]{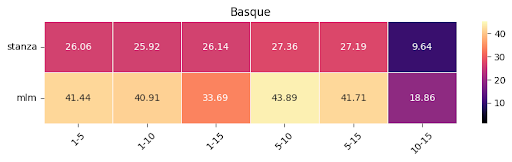}     
    \caption{Grid search F1 scores for scoring-based methods MLM and Stanza. The x axis indicates minimum and maximum number of words of the segmentation window.}
    \label{fig:gridsearch}
\end{figure}

The complete results of the grid-search described and discussed in Section~\ref{sec:err_analysis} are shown in Figure~\ref{fig:gridsearch}.

\begin{figure*}[ht]
\centering
\includegraphics[scale=0.8]{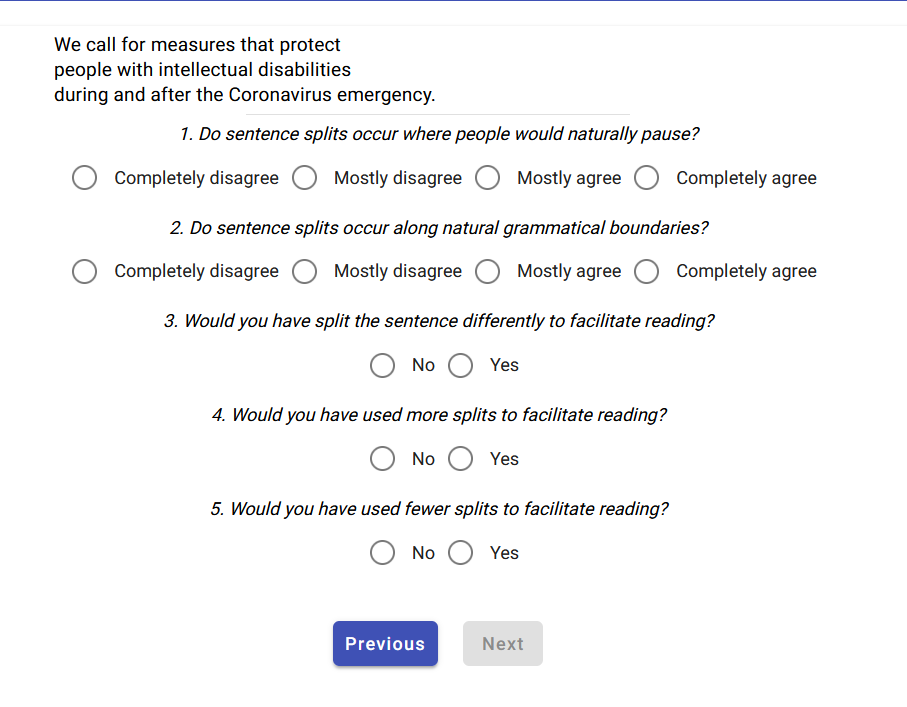}
\caption{Web interface for the qualitative evaluation}
\label{fig:qualweb-eval}
\end{figure*}

\section{Human Evaluation Protocol}
\label{sec:appendix-heval}

To evaluate system hypotheses, there were 3 evaluators per language; for human segmentation, there were 6, 4 and 3 evaluators for English, Spanish and Basque, respectively. All evaluators were unpaid volunteers, proficient in the corresponding language. 

The samples and associated questions were provided within an ad-hoc Web-based environment developed for the task, shown in Figure~\ref{fig:qualweb-eval}. The following instructions were provided to the evaluators:

The general guidelines to facilitate reading via segmentation are as follows:

\begin{itemize}
    \item Insert line breaks (splits) that follow natural linguistic boundaries.
    
    \item  Try to ensure that each line has a minimum of 5 words and a maximum of 15. This recommendation is not mandatory and it is advised to prioritise what would make reading easier.    
\end{itemize}

The meaning of each question in the evaluation is indicated below and guidelines are provided to respond appropriately:

\begin{enumerate}
    \item Do the splits in the sentence occur where a person would pause when reading it?

    \begin{itemize}

        \item This question aims to assess whether the splits correspond to positions in the sentence where it would be natural for a native speaker to pause while reading.

        \item Example of a natural split as a pause:

        \begin{itemize}
            \item People participated in the experiment \\ with enthusiasm during the day.
        \end{itemize}

        \item Example of a less natural split as a pause:

        \begin{itemize}
            \item People participated in the \\ experiment with enthusiasm during the day.
        \end{itemize}

        \item In the event that there is more than one break in the sentence:

        \begin{itemize}

            \item If all splits are natural, indicate: Totally agree.
            \item If the majority of splits are natural, indicate: Partially agree.
            \item If the majority or half the splits are unnatural, indicate: Partially disagree.
            \item If none is natural, indicate: Totally disagree.
        \end{itemize}
        
    \end{itemize}
    
    \item Do the splits in the sentence preserve linguistic boundaries?

    \begin{itemize}

    \item This question aims to assess whether the splits preserve grammatical boundaries.

        \item Example of a natural split in terms of grammatical boundaries:

        \begin{itemize}
            \item People participated \\ 
            in the experiment \\ with enthusiasm \\
            during the day.
        \end{itemize}

        \item Example of a less natural split in terms of grammatical boundaries:

        \begin{itemize}
            \item People \\ participated in the \\ experiment with 
            \\enthusiasm during the day.
        \end{itemize}

        \item In the event that there is more than one split in the sentence:

        \begin{itemize}

            \item If all splits preserve grammatical boundaries, indicate: Totally agree.
            \item If the majority of splits preserve grammatical boundaries, indicate: Partially agree.
            \item If the majority or half the splits do not preserve grammatical boundaries, indicate: Partially disagree.
            \item If none preserve grammatical boundaries, indicate: Totally disagree.
        \end{itemize}    
    
    \end{itemize}

    \item  Would you have used different splits in the sentence to make it easier to read?

    \begin{itemize}
        \item Answer YES if you would have modified the splits in any way:
        
        \begin{itemize}
            \item Moving the current splits to another position and/or
            \item Adding splits and/or
            \item Removing splits.
        \end{itemize}

        \item Answer NO if you would not make any change to the current splits.

    \end{itemize}

    \item Would you have used more splits in the sentence to make it easier to read?

    \begin{itemize}
        \item Answer YES if you had used more splits.
        \item Answer NO otherwise.        
    \end{itemize}

    \item Would you have used fewer splits in the sentence to make it easier to read?

    \begin{itemize}
        \item Answer YES if you had used fewer splits.
        \item Answer NO otherwise.      
    \end{itemize}

\end{enumerate}

\section{Segmentation Examples}
\label{sec:appendix-examples}

Tables~\ref{table:exples-mlm-good},~\ref{table:exples-mlm-bad} and~\ref{table:exples-human-bad} provide examples of correct  MLM-scoring segmentation, incorrect MLM-scoring segmentation, and incorrect human segmentation, respectively.

\begin{table*}[ht]
\centering
\begin{tabular}{ll}
\toprule

Language                & Segmentation    \\ \midrule

\multirow{6}{*}{EN} 
& \begin{tabular}[c]{@{}l@{}}It means that people with disabilities \\ could not have access to healthcare.\end{tabular}  \\  \cmidrule(lr){2-2}
                    
& \begin{tabular}[c]{@{}l@{}}In order to find a job, \\ we also need to access vocational training.\end{tabular}  \\ \cmidrule(lr){2-2}
                    
& \begin{tabular}[c]{@{}l@{}}To this date, the NGO has about 60 volunteers,\\ who help about 60 children with disabilities\end{tabular} \\

\midrule

\multirow{10}{*}{ES} 

& \begin{tabular}[c]{@{}l@{}}También se han debatido las enmiendas\\ que han realizado los demás partidos políticos.\\ \textit{The amendments that were made by the other political parties were also debated there.}\end{tabular} \\ \cmidrule(lr){2-2}

& \begin{tabular}[c]{@{}l@{}}En los cursos que se hacen en la Escuela de Pastores\\ la cuarta parte de los alumnos han sido mujeres\\ y esa presencia es importante y hay que hacerla visible.\\\textit{In the courses that are held at the Escuela de Pastores the fourth part of the students there} \\ \textit{have been women and that presence is important and has to be made visible.}\end{tabular} \\ \cmidrule(lr){2-2}

& \begin{tabular}[c]{@{}l@{}}La lectura fácil es útil para muchas personas\\ con discapacidad intelectual o del desarrollo.  \\ \textit{Easy reading is useful for many people with intellectual or developmental disabilities.}  \end{tabular}    \\ 

\midrule

\multirow{8}{*}{EU} 

& \begin{tabular}[c]{@{}l@{}}Horrela, pertsonek elkarrizketa eta akordioak egiten dituzte,\\ bata besteen artean aurka egin ordez.  \\ \textit{This way, people make conversations and agreements, instead of opposing each other.} \end{tabular} \\ \cmidrule(lr){2-2}

& \begin{tabular}[c]{@{}l@{}}Bilera honekin indartu nahi da\\ bi erakundeen arteko lankidetza.   \\ \textit{This meeting is intended to strengthen collaboration between the two organizations.}  \end{tabular}  \\ \cmidrule(lr){2-2}

& \begin{tabular}[c]{@{}l@{}}Sistema horrek herritarrei abisua ematen die\\ larrialdi handi edo hondamendi kasuetan.\\ \textit{This system warns citizens in case of major emergency or disaster.} \end{tabular}  \\ 
\bottomrule
\end{tabular}
\caption{Examples of MLM scoring-based segmentation viewed as correct}
\label{table:exples-mlm-good}
\end{table*}

\begin{table*}[ht]
\centering
\begin{tabular}{ll}
\toprule

Language                & Segmentation    \\ \midrule

\multirow{6}{*}{EN} 
& \begin{tabular}[c]{@{}l@{}}We talked about deinstitutionalisation with fifteen \\countries of the European Union.\end{tabular}  \\  \cmidrule(lr){2-2}
                    
& \begin{tabular}[c]{@{}l@{}}People working in institutions did \\ not have enough protections (masks, gloves).\end{tabular}  \\ \cmidrule(lr){2-2}
                    
& \begin{tabular}[c]{@{}l@{}}EPSA members talked about asking for more \\ money to solve this problem.\end{tabular} \\

\midrule

\multirow{10}{*}{ES} 

& \begin{tabular}[c]{@{}l@{}}Un delito por ejemplo puede \\ ser que te hayan pegado.\\ \textit{A crime, for example, could be that you have been hit.}\end{tabular} \\ \cmidrule(lr){2-2}

& \begin{tabular}[c]{@{}l@{}}Si necesitas más información, te \\ recomendamos leer el reglamento original. \\\textit{If you need more information, we recommend reading the original regulations.} \end{tabular} \\ \cmidrule(lr){2-2}

& \begin{tabular}[c]{@{}l@{}}Pero que es más importante \\ aún, que los cocineros y cocineras del jurado \\ se hayan comprometido con su proyecto Zero Foodprint.\\ \textit{But what is even more important is that the chefs on the jury } \\ \textit{have committed to their Zero Foodprint project.} \end{tabular}    \\ 

\midrule

\multirow{8}{*}{EU} 

& \begin{tabular}[c]{@{}l@{}}Azkenik, sailburuak esan du istripuarekin lotutako prozesu \\ administratibo eta judizial guztiek aurrera jarraitzen dutela.  \\ \textit{Finally, the minister said that all administrative and judicial processes} \\ \textit{related to the accident proceed.} \end{tabular} \\ \cmidrule(lr){2-2}

& \begin{tabular}[c]{@{}l@{}}Dirua gehiagotu da Zinematografiaren eta Arteen Institutuarekin \\ (ICCA) hitzarmen bat sinatuko delako.  \\ \textit{The money has increased because an agreement will be signed } \\ \textit{with the Institute of Cinematography and Arts (ICCA).}  \end{tabular}  \\ \cmidrule(lr){2-2}

& \begin{tabular}[c]{@{}l@{}}Euskadik ez du ahaztuko izan zuten gizatasuna \\ eta elkartasuna.\\ \textit{Euskadi will not forget their humanity and solidarity.} \end{tabular}  \\ 
\bottomrule
\end{tabular}
\caption{Examples of  MLM scoring-based segmentation viewed as incorrect}
\label{table:exples-mlm-bad}
\end{table*}

\begin{table*}[ht]
\centering
\begin{tabular}{ll}
\toprule

Language                & Segmentation    \\ \midrule

\multirow{6}{*}{EN} 
& \begin{tabular}[c]{@{}l@{}}Atempo is a company that \\ helps people with disabilities.\end{tabular}  \\  \cmidrule(lr){2-2}
                    
& \begin{tabular}[c]{@{}l@{}}The conference was about \\ the 10 years of deinstitutionalization, in Europe.\end{tabular}  \\ \cmidrule(lr){2-2}
                    
& \begin{tabular}[c]{@{}l@{}}Paul Alford did not have a \\ choice on whom to live with.\end{tabular} \\

\midrule

\multirow{10}{*}{ES} 

& \begin{tabular}[c]{@{}l@{}}Conoce cómo funciona el Parlamento \\ de La Rioja, su historia \\ y curiosidades.\\\textit{Learn how the Parliament of La Rioja works, its history and curiosities.}\end{tabular} \\ \cmidrule(lr){2-2}

& \begin{tabular}[c]{@{}l@{}}Eso quiere decir que es \\ fácil de entender. \\ \textit{That means it is easy to understand.} \end{tabular} \\ \cmidrule(lr){2-2}

& \begin{tabular}[c]{@{}l@{}}Erkoreka ha dicho que el acuerdo es posible \\ pero es necesario que el Gobierno español también \\ avance y cumpla con los compromisos que han tomado. \\ \textit{Erkoreka has said that the agreement is possible but it is necessary for the Spanish} \\ \textit{Government to also move forward and fulfill the commitments they have made.} \end{tabular}    \\ 

\midrule

\multirow{8}{*}{EU}

& \begin{tabular}[c]{@{}l@{}}Sektore horiek ere lotuta \\ daudelako klimarekin. \\ \textit{Because these sectors are also linked with the climate.} \end{tabular} \\ \cmidrule(lr){2-2}

& \begin{tabular}[c]{@{}l@{}}Bonoak erosten dituzten pertsonek 2021eko maiatzaren 31ra arte erabili ahal \\ izango dituzte. \\ \textit{People who buy vouchers will be able to use them until May 31, 2021.} \end{tabular}  \\ \cmidrule(lr){2-2}

& \begin{tabular}[c]{@{}l@{}}Guztira, miloi erditik gora euro emango \\ dira, lau diru-laguntzatan banatuta. \\ \textit{In total, more than half a million euros will be given, divided into four grants.}\end{tabular}  \\ 
\bottomrule
\end{tabular}
\caption{Examples of human references viewed as incorrect}
\label{table:exples-human-bad}
\end{table*}

\end{document}